# CLASSIFICATION OF THINGS IN DBPEDIA USING DEEP NEURAL NETWORKS

Rahul Parundekar

Elevate.do
`rahul@elevate.do`

*ABSTRACT*

*The Semantic Web aims at representing knowledge about the real world at web scale - things, their attributes and relationships among them can be represented as nodes and edges in an inter-linked semantic graph. In the presence of noisy data, as is typical of data on the Semantic Web, a software Agent needs to be able to robustly infer one or more associated actionable classes for the individuals in order to act automatically on it. We model this problem as a multi-label classification task where we want to robustly identify types of the individuals in a semantic graph such as DBpedia, which we use as an exemplary dataset on the Semantic Web. Our approach first extracts multiple features for the individuals using random walks and then performs multi-label classification using fully-connected Neural Networks. Through systematic exploration and experimentation, we identify the effect of hyper-parameters of the feature extraction and the fully-connected Neural Network structure on the classification performance. Our final results show that our method performs better than state-of-the-art inferencing systems like SDtype and SLCN, from which we can conclude that random-walk-based feature extraction of individuals and their multi-label classification using Deep Neural Networks is a promising alternative to these systems for type classification of individuals on the Semantic Web. The main contribution of our work is to introduce a novel approach that allows us to use Deep Neural Networks to identify types of individuals in a noisy semantic graph by extracting features using random walks.*

## 1. INTRODUCTION

The world around us contains different types of things (e.g. people, objects, ideas, etc.) that have attributes (e.g. shape, color, etc.) and relationships with other things. For example, Washington, D.C. is a city (like Paris, Berlin, etc.) with attributes like population, location, etc. while U.S.A is a country (like France, Germany, etc.) with its own set of attributes. Washington D.C. also has a relationship with U.S.A. - that of being its capital, which adds extra meaning to it. Semantic Graphs are well suited for representing such conceptual knowledge about the world. A software Agent operating on such data can then reason about the world and perform actions to satisfy system goals (e.g. user commands). The Semantic Web [1] and Linked Data Web [2] aim at allowing Agents to reason and act at web-scale using standardized knowledge representation languages and services. Semantic Graphs are also used in other domains like Spoken Dialog Systems [3], Social Networks [4], Scene Understanding [3], Virtual & Augmented Reality [5], etc.

Traditionally, software Agents process the information in such Semantic Graphs using inference rules to infer actionable classes. For example, the Semantic Web's OWL-DL inferencing uses SHIQ description logics [6] for inferences. These inferred classes can then be used to actuate actions by executing stored procedures. Since the Agent is only as powerful as the ability of its inference rules to infer the actionable classes, its capability is limited by the expressive power of the inference mechanism and the developers' ability to create appropriate inference rules. In addition, as pointed out in Paulheim et. al [7], Semantic Web data in the wild contains a lot of noise. Even a single noisy instance along with the inference rules can break the entailments in the classes and add new entailments that may be incorrect. The field of Machine Learning offers a solution - if an Agent is able to learn to classify the semantic data into actionable classes by understanding the attributes and relationships of things instead of using manually defined rules, it can perform the inference more robustly and overcome the issues mentioned above. Corresponding to the domains listed in the previous paragraph, applications of such an Agent could be a Semantic Search Engine [8], a virtual assistant [9] [10], a recommender [4], an autonomous situation understanding and responsive system [11], etc.

To achieve robust inferencing, we create an approach that extracts features for individuals from a Semantic Graph and performs multi-label classification to identify the types of those individuals. We conducted a prior art search of existing solutions (see Section 2) and decided to classify individuals in a well-known Semantic Web dataset called DBpedia (see Section 3) for effective comparison of the performance of our approach. Our approach consists of two parts. First, we extract features for the individuals being classified using a random walk technique (see Section 4). Second, we train a multi-label classifier to learn to estimate the classes on a part of the data (training set) and use it to classify unseen data (see Section 5). We use a Deep Learning approach for the multi-label classification due to the large yet sparse features extracted by the random walk approach. Since there is no prior art exploring this combination to the best of our knowledge, we systematically explore the effectiveness of different hyper-parameters of our random walk algorithm and Neural Network structures to perform the multi-label classification. After selecting the best hyper-parameters, we perform classification for our chosen dataset and present the results by comparing it with other approaches (see Section 6). From the results, we conclude (see Section 7) that random walk-based feature extraction and multi-label classification using Deep Neural Networks is a promising approach to understanding things and contexts robustly in a Semantic Graph.

## 2. PRIOR ART

There has been some prior art in using Machine Learning to infer the types of individuals in Semantic Web data [12] [13]. Most relevant of these are the SDtype [7] and SLCN [14], which use only the incoming relationships of an individual to infer its types. For individuals in DBpedia, SDtype uses a heuristic link-based inference mechanism and achieves an $F_1$-score (a classifier's $F_1$-score is the harmonic mean of its precision & recall) of 0.765 for identifying DBpedia Ontology types and a score of 0.666 for identifying Yago types. SLCN creates local classifiers for each node in the type hierarchy and then uses a top-down approach to infer the type of classes at lower levels. In comparison, it achieves a score of 0.847 for DBpedia types and 0.702 for Yago types. MLC4.5 [15] is another approach that uses a modified C4.5 decision tree algorithm and learns to classify types of all the classes in the hierarchy at once.

These approaches have some limitations. MLC4.5 is not able to scale for large datasets such as DBpedia, SDtype only works on specific type of features [14], and SLCN relies on the presence of a well-defined hierarchical structure, which may be absent if the Ontology defined for the sources is rudimentary. To use Machine Learning, we need to extract features from the Semantic Graph for the individuals to make the data classifier-ready. Graph Kernels can be used to extract such features [16]. We selected a random walk approach for its simplicity and ability to extract neighbourhood context (see Section 4). We use Deep Learning to learn the complex class definitions and create a Multi-label classifier [17], since recent work has shown their effectiveness in similar domains [18]. There has also been some work in combining random walks on graphs with Deep Learning. Perozzi et. al [19], for example, use a social network graph for link prediction and recommendation. It is, however, a social graph and NOT a semantic graph. After considering all the datasets used in the prior art, we decided to use DBpedia as an exemplary large-scale Semantic Graph to study type inferencing for effective comparison of the performance of our approach with those in the prior art.

## 3. USING THE DBPEDIA DATASET

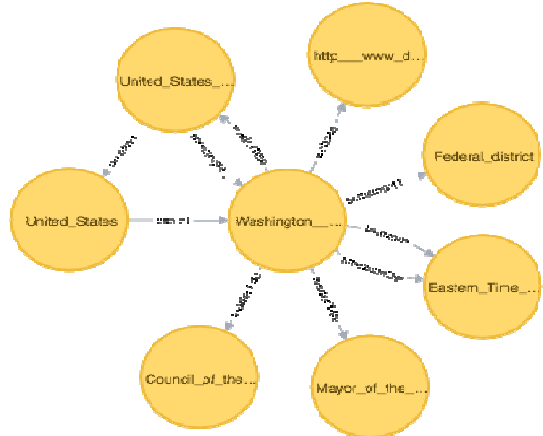

Figure 1. Example neighbourhood of Washington D.C. in the DBpedia Semantic Graph and its relationship to United States.

DBpedia is an encyclopaedic graph of structured information extracted from Wikipedia info-boxes, page links, categories, etc. [20]. For our earlier example, we can see U.S.A. (the 'United States' node on the left) connected to Washington D.C. using the capital relationship in the subset of the Semantic Graph in DBpedia shown in Figure 1. Each thing in DBpedia has one or more types and categories associated with it. While the user community creating DBpedia maintains an Ontology, which specifies the types hierarchy of things (e.g. the types associated with U.S.A. and Washington D.C. can be seen in Figure 2), it additionally specifies the types of things in relation to the Yago [21] dataset (among others) as well as extracted categories for them. We use the subset of DBpedia (*http://wiki.dbpedia.org/downloads-2016-04*), which was generated from the March/April 2016 dump of Wikipedia (Note: SDtype and SLCN use the 2014 DBpedia dumps). We classify the individuals in this dataset using their attributes and their relationships extracted from Wikipedia info-boxes (from *infobox_properties_en.ttl* having 30 million triples) into classes from three different hierarchies - the DBpedia Ontology types (from *instance_types_en.ttl* having 5.3 million triples), the Yago Ontology types (from *yago_types.ttl* having 57.9 million triples) and the categories (*article_categories_en.ttl* having 22.5 million triples) they belong to. We only use the English dumps in the N-Triple format [20].

**Examples Types for United_States:**
Settlement
Country
PopulatedPlace
...

**Examples for Washington__D_C_:**
Place
Settlement
schema.org:Place
...

Figure 2. Examples of Types in the DBpedia-OntologyType Dataset.

## 3. FEATURE EXTRACTION USING RANDOM WALKS

To train our multi-label classifier, we need to extract a set of features from individuals in the Semantic Graph. While using only the attributes of the individuals as features is trivial as noted by Paulheim et. al [7], we decide to generate features that also capture the relationships of the individuals with others in their neighbourhood. This is motivated by the Washington D.C. and the U.S.A example mentioned earlier where the attributes of these and the relationship between the two gives an additional meaning of the city being the "capital" of U.S.A.

Our random walk approach [16] to extract features for each individual is as follows. We start our walk on the individual for which the features are to be extracted. We then create a list of possible steps that we can take from that node from one or more different types of steps available. With DBpedia, the four type of steps available are - presence of an attribute, presence of an outgoing relationship, presence of an incoming relationship (for all these three, after taking the step we will stay on the same node), and a step over an incoming or outgoing relationship (here, we will land on a different node with whom the node has the relationship). We then select one step from this list of available steps randomly to land on the next node. We repeat this for $l$ such steps. We use the names of the attributes and relationships we observe while taking the $l$ sequential steps to generate a list of ordered labels. This ordered list of labels generated from our random walk of length $l$ then becomes one feature for our classification. By performing $n$ such random walks starting at the individual to be classified, we can extract up to $n$ distinct features for that individual. We repeat this for each individual, to extract features for the entire dataset.

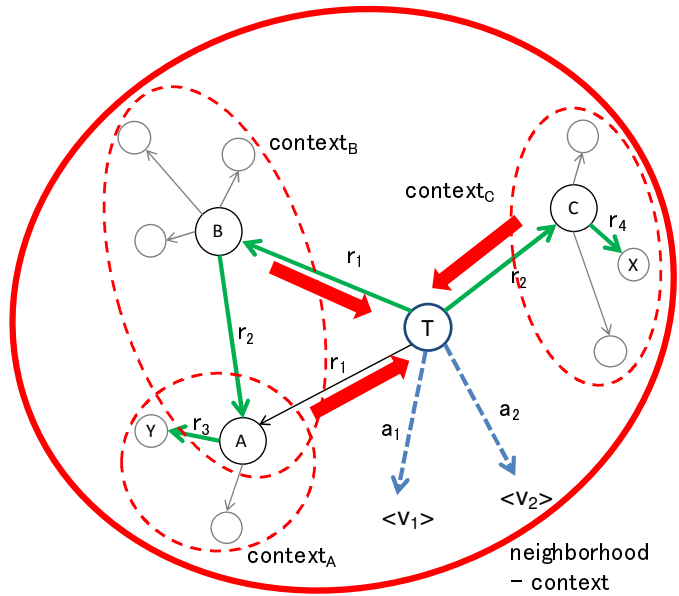

Figure 3. Neighborhood Context Extracted with Random Walks

As shown in Figure 3, features generated using the random walk approach aim to capture neighbourhood contexts using other individuals around the individual to be classified. The simplest random walk we can take is of length 1. Starting at node T, we can choose a 'step' that notes the attributes present and stay on T. In this case, the feature extracted signifies the presence of an attribute - e.g. '$a_1$'. A random walk of length 2 using only such attribute presence step generates a complex feature '$a_1, a_2$' (the comma signifies conjunction). Another random walk of length 2 may go to C and its neighbour X as seen on the right in the figure. The feature extracted

International Journal of Web & Semantic Technology (IJWesT) Vol.9, No.1, January 2018here would be '$r_2$->$r_4$' (the arrow signifies 'moving'). A feature extracted with a random walk of length 3 can be '$r_1$->$r_2$->$r_3$' as the walk moves from T to B, A and finally ending at Y. Features generated for the individual in such a manner are a combination of its attributes, the relationship it participates in and the attributes and relationships of its neighbours, and their neighbours. When we take multiple walks within this neighbourhood, the context of the neighbourhood starts adding up as shown by the block arrows in the figure.

### 4.1. Pre-processing

Our target classes for each individual can come from three files mentioned earlier - types from the DBpedia Ontology, the DBpedia categories, or types from the Yago Ontology. To extract the multi-label target classes, we first take the inner join of each of those three sources with the individuals available in the graph. We then create a multi-label target vector for each individual such that the value for the label for a type *t* in the vector is 1 if the individual is an instance of type *t*, or else is 0. Once we identify the inner-join and the multi-label vectors, we remove the individuals from the Semantic graph that do not have type information. While this step reduces the possible relationships that we may encounter with individuals for whom there is no type information available, it also removes spurious individuals (e.g. things generated from Wikipedia articles that have no distinguishing attributes, things in external datasets for which we do not have type data, etc.). We end up with three reduced datasets - DBpedia-OntologyTypes, DBpedia-Categories, DBpedia-YagoTypes, the statistics about which are described in Table 1.

Table 1. Statistics for the Three Reduced Datasets.

| Semantic Graph | Number of (#) individuals | Average # attributes | Average # incoming relationships | Average # relationships | # Types / Categories |
|---|---|---|---|---|---|
| **DBpedia-OntologyTypes** | 3.18M | 5.78 | 3.58 | 2.74 | 526 |
| **DBpedia-Categories** | 2.26M | 6.25 | 2.89 | 2.11 | 1099 |
| **DBpedia-YagoTypes** | 2.78M | 5.97 | 4.21 | 3.08 | 2083 |

### 4.2. Random Walk Hyper-parameters

We have three hyper-parameters for our random walk algorithm that affect the kind of features that we can extract for the individual to be classified. Choosing different values for these results in different kinds of features we can extract from our random walks algorithm.

1) Step Strategy: In performing our random walk, we have three step selection strategies.

   a. The Stay step strategy has 3 kinds of steps we can take:

      i. Attribute presence: where we pick randomly among the attributes present, note the attribute name, and stay on the same node (see Figure 4 for examples).

      ii. Relationship presence: where we pick randomly among the relationships present, note the relationship name, and stay on the same node (see Figure 5 for examples).

      iii. Incoming Relationship presence: where we pick randomly among the incoming relationships present, note the incoming relationship name, and stay on the same node (see Figure 6 for examples).

    b. The Move step strategy contains only one kind:

        i. Relationship Step: where we pick randomly among the incoming and outgoing relationships and move to the node connected to by the relationship.

    c. The Both step strategy that is a combination of the two.

In the strategies above, we do not use the values of the attributes or the identifiers of the individual in this work. While this is possible, additional systematic research is needed to understand binning strategies and removal of features generated that act as secondary keys as a result of one-to-one or one-to-many relationships. This also allows us to compare with the prior art, which similarly ignores such values.

---

**Examples for United_States:**
has_imageFlag
has_ethnicGroups
has_populationEstimate
...

**Examples for Washington__D_C_:**
has_elevationMinFt
has_populationRank
has_areaCode
...

---

Figure 4. Examples of Labels of Attribute Presence Steps.

---

**Examples for United_States:**
hasRel_demonym
hasRel_capital
hasRel_largestCity
...

**Examples for Washington__D_C_:**
hasRel_leaderTitle
…

---

Figure 5. Examples of Labels of Relationship Presence Steps.

---

**Examples for United_States:**
hasInRel_almaMater
hasInRel_channel
hasInRel_based
...

**Examples for Washington__D_C_:**
hasInRel_regions
hasInRel_origin
hasInRel_restingPlace
...

---

Figure 6. Examples of Labels of Relationship Presence Steps.

2) Length of the Walk: The length hyper-parameter specifies the number of steps we take to extract one feature. In the "Stay" step strategy above, a longer length means that we create one combined single feature based on the co-occurrence of the presence of the attributes and relationships that the individual has. In the "Move" step strategy, a longer length creates a combined feature that encodes neighbour relations. The "Both" step strategy creates features that are a combination of both above, thus extracting the neighbourhood context as shown in Figure 3.

3) Length Strategy: We also use two strategies for selecting the length - a fixed length of exactly 2 or 3, and a variable length of up to 2 or 3 with a higher probability for walks with lower lengths (see comments in Figure 7 for details).

Before we go any further, it is important to note that the semantics of the binary values (either 0 or 1) of the features extracted are different than classic Machine Learning. In classic Machine Learning, the feature value is 0 if the feature is absent or is 1 if it is present. With the random walk method, however, the meaning changes. The feature value is 1 when a random walk is observed, but it is 0 when the random walk is not observed even if it may be present.

### 4.3. Algorithm

Figure 7 shows the algorithm for extracting features using random walks. The inputs to the function are - the Semantic Graph, the number of features to be extracted (n), the maximum length of the walk, and the type of steps allowed (i.e. steps from the Stay and Move categories). The function returns the walks extracted for the individual. For the sake of brevity, the details of the algorithm are explained in-line in the code. We take some additional precautions in the above algorithm while extracting features. We make sure that the walks are not empty, which could happen if there are no attributes or relationships extracted for the individual from Wikipedia, and the individuals for such cases are removed from the dataset. Also, since the next node after taking the steps from the Stay category is the same node, it may introduce multiple features with the same set of labels but in a different order. To eliminate this inconsistency, we sort the steps taken at the same node in a lexicographic order before adding their sequence to the walk. It should be noted that even if we ask our algorithm to return $n$ number of walks the total number of features extracted may not be $n$ if some of the walks are observed multiple times. Finally, we split the data into training and testing datasets. The datasets were first split into batches of up to 5000 individuals in each (lesser, if unnecessary individuals were removed). The batches were then split into 80% training - 20% testing split.

```
def extractFeatures(semanticGraph, n, maxLength, stepTypes):
  walkIndex = {}
  walks = {}
  for individual in semanticGraph:
    # For n number of walks
    for i in range(n):
      # To choose the length, we compare 2 strategies:
      # Strategy 1: Fixed Length
      #   Here, the length is fixed to maxLength
      # Strategy 2: Variable length from 1 upto maxLength
      #   Here, the length is chosen with probability
      #     (maxLength-l+1) / (1+2+...+maxLength)
      #   This allows shorter walks to be more dominant.
      #   For example, chooseLength(2)
      #     returns l=1 with probability 2/3
      #     returns l=2 with probability 1/3
      l = chooseLength(maxLength)
```

```
      walk = []
      currentNode=individual
      for step in range(l):
        availableSteps = []
        nextNodes = []
        # Create list of available steps from stepTypes
        # (Assume standard graph helper functions below)
        # 1. Attribute presence
        if (`attribute' in stepTypes):
          for attr,value in currentNode.attributes():
           if((`hasAttr_' + attr) not in availableSteps):
             availableSteps.append(`hasAttr_' + attr)
             nextNode.append(currentNode)
        # 2. Relationship presence
        if (`relationship' in stepTypes):
          for rel, node in currentNode.links():
           if((`hasRel_' + rel) not in availableSteps):
             availableSteps.append(`hasRel_' + rel)
             nextNode.append(currentNode)
        # 3. Incoming relationship presence
        if (`incoming' in stepTypes):
          for rel, node in currentNode.incomingLinks():
           if((`hasInRel_' + rel) not in availableSteps):
             availableSteps.append(`hasInRel_' + rel)
             nextNode.append(currentNode)
        # 4. Relationship step
        # If l=1, then this is same as 2 and 3,
        # and so we add these only if l>1
        if (`step' in stepTypes && l>1):
           for rel, node in currentNode.links():
              availableSteps.append(rel + `->')
              nextNode.append(node)
           for rel, node in currentNode.incomingLinks():
              availableSteps.append (rel + `<-')
              nextNode.append(node)
        # Take a step randomly
        stepIndex = randInt(0,len(availableSteps)-1)
        step = availableSteps[stepIndex]
        nextNode = nextNodes[stepIndex]
        walk.append(step) #Append to the walk
        currentNode = nextNode #Move to next node
    #Add to the walks
    if walk not in walks:
        walks.append(walk)
return walks
```

Figure 7. Algorithm for Feature Extraction Using Random Walks

## 5. CLASSIFICATION USING DEEP NEURAL NETWORKS

We use Deep Neural Networks for the multi-label classification. With the lack of significant prior work, we systematically explore the relationship between the classification performance of different fully-connected Neural Network structures and the features extracted using the random walk approach to study their effectiveness for our dataset.

### 5.1. Performance Vis-à-vis Network Structure and Kind of Steps

The simplest case we study is the relationship between depth of the fully-connected Neural Network and the features generated using random walks of length 1. We select three network structures with varying depth for this case.

1. **Logistic Regression:** The simple logistic regression network has only one fully-connected layer. This will help us establish a baseline with respect to the other two structures and provide us a sanity check. Figure 8 shows the simple network architecture.

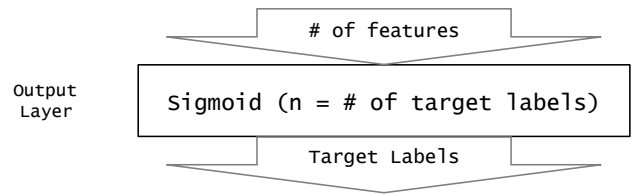

Figure 8. Simple Logistic Regression Network Structure

2. **4 Layer Fully-connected Neural Network:** The first deep fully-connected Neural Network contains 4 layers including the input and output. The shape and sizes of the layers can be seen in Figure 9 (Note: similar layers collapsed for simplicity).

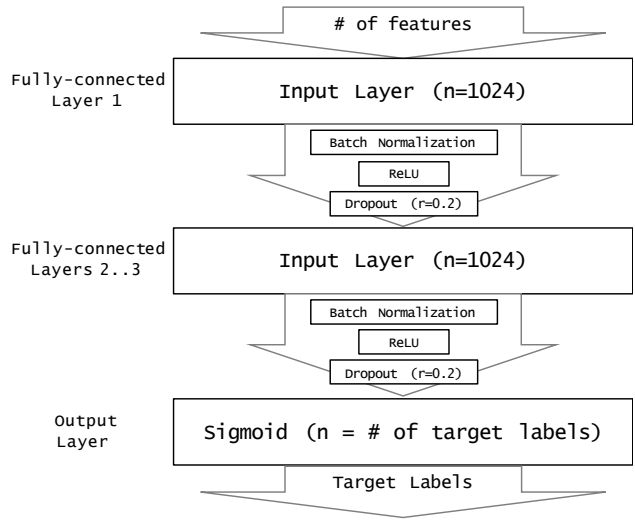

Figure 9. 4 Layer Deep Neural Network Structure

3. **8 Layer Fully-connected Neural Network:** The second deep fully-connected Neural Network contains 8 layers including the input and output. Compared to the 4-layer network, we chose to explore fewer neurons per layer but deeper structure. The shape and sizes of the layers can be seen in Figure 10.

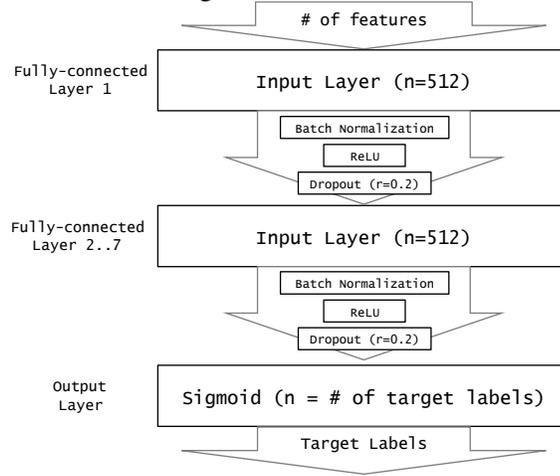

Figure 10. 8 Layer Deep Neural Network Structure

We use the *extractFeatures* function to extract features of the stay category of steps. We select incrementing number of 5, 10, 25, and all available features for our random walks and explore the ability of the 3 structures described above to learn to classify the reduced DBpedia-OntologyTypes dataset. This dataset has a total of 3.18M individuals and 529 target classes. In the three Neural Network structures above, the features extracted become inputs to the first layer and the target classes are outputs of the final layer. We used Xavier initialization (i.e. Glorot Normal initialization [22]) to initialize the weights of neurons as it avoids extreme values of the weights that start in dead or saturated regions and instead lets the signal reach deeper into the network. Based on some preliminary exploration, we trained each for 5 epochs as the $F_1$-score plateaus or, in the case of the 8-layer network, starts overfitting after that. For learning, we use Adam optimizer with the Nesterov accelerated gradient approach [23], since it is an adaptive gradient method and our data is sparse. We use 'binary cross-entropy' loss function for learning. Except for the Simple Logistic Regression classifier, we have the following additional structural features. We use Batch Normalization for accelerated learning. We use a Dropout layer to avoid overfitting, especially since the data is sparse and the network may learn highly-specialized definitions. We chose a dropout rate of 0.2 after experimentations showed that a rate of 0.4 resulted in slow convergence without improvement in performance. Other hyper-parameters (e.g. learning rate) were left to default values in Keras (https://github.com/fchollet/keras).

Tables II, III, IV, and V show the $F_1$-scores of the three structures for a different number of walks of length one and using attribute presence, relationship presence, incoming relationships, and all the three kinds of steps. We can see that among the three types of steps, the attribute presence step by itself has the most information to identify the classes. Meanwhile, the maximum $F_1$-score we get is 0.9235 when we use all three step-types. The incoming relationship presence step scores the least at 0.7597. We note that performance increases with more random walks and the fully-connected Neural Network with 4 layers performs the best in all cases.

Table 2. Performance using attribute presence step only.

| # of random walks | # distinct features | Logistic Regression | 4 Layers Fully Connected | 8 Layers Fully Connected |
|---|---|---|---|---|
| **10** | 1993 | 0.8224 | 0.8596 | 0.8590 |
| **15** | 2006 | 0.8330 | 0.8687 | 0.8683 |

| 20 | 2014 | 0.8374 | 0.8733 | 0.8727 |
| All | 2020 | 0.8421 | 0.8765 | 0.8767 |

Table 3. Performance using relationship presence step only.

| # of random walks | # distinct features | Logistic Regression | 4 Layers Fully Connected | 8 Layers Fully Connected |
|---|---|---|---|---|
| 10 | 1066 | 0.7720 | 0.7956 | 0.7995 |
| 15 | 1072 | 0.7751 | 0.8001 | 0.7981 |
| 20 | 1071 | 0.7749 | 0.8058 | 0.8007 |
| All | 1075 | 0.7743 | 0.8025 | 0.7989 |

Table 4. Performance using incoming relationship presence step only.

| # of random walks | # distinct features | Logistic Regression | 4 Layers Fully Connected | 8 Layers Fully Connected |
|---|---|---|---|---|
| 10 | 1043 | 0.6557 | 0.7619 | 0.7612 |
| 15 | 1072 | 0.6596 | 0.7597 | 0.7586 |
| 20 | 1071 | 0.6603 | 0.7563 | 0.7629 |
| All | 1075 | 0.6693 | 0.7476 | 0.7596 |

Table 5. Performance using all 3 step-types.

| # of random walks | # distinct features | Logistic Regression | 4 Layers Fully Connected | 8 Layers Fully Connected |
|---|---|---|---|---|
| 10 | 4014 | 0.8256 | 0.8987 | 0.8960 |
| 15 | 4092 | 0.8428 | 0.9097 | 0.9083 |
| 20 | 4110 | 0.8484 | 0.9158 | 0.9144 |
| All | 4196 | 0.8613 | 0.9235 | 0.9219 |

### 5.2. Performance Vis-à-vis Random Walk Hyper-parameters

The next case we study is the effect of the selection of hyper-parameters of the random walk algorithm - step strategy, length of the walks, and length strategy. We started with the Neural Network structure with 4 Layers (see Figure 9) as our starting point for this phase based on the conclusions of the previous case and tried out the following improvements before settling down on a similar structure, but with 6 layers:

- We increased the depth as much as we could, but noticed that after 6 layers the network starts overfitting the data as evidenced by the divergence between the training and testing $F_1$-score after initial simultaneous increase.

- We tried layers with 2048 neurons as well. However, we noted that as we start random walks of length longer than one step, our number of features grows very rapidly and even goes into millions. With feature vectors of that length, given the memory of the GPU (4GB), we restricted to 1024 neurons to not run into memory overflow.

- We also increased the number of walks we perform to 25, 50 and 100 to extract more features since our walk length has increased.

We soon realized that even with 1024 neurons in the input layer, we quickly ran out of memory while training our model due to the extremely large number of features extracted by the random walk algorithm (e.g. with 125 random walks of length 2 taken for individuals, the number of distinct features extracted is in the order of hundreds of thousands). So, we encode our inputs into a feasible scheme. Since the dataset is sparse and we may only have 100 features in each row at maximum, we decided to superimpose features. We select an input feature size of 8384 neurons (1024x8). If the ith feature is present, our encoding function is:

$$X[i \mod 8384] = \lfloor i \div 8384 \rfloor + 1$$

We thus have 36 variations of our hyper-parameters for our next experiments:

- Length of walks - 2 or 3
- Number of walks - 25, 50 or 100
- Category of steps - Stay (see Figure 11), Move (see Figure 12), or Both (see Figure 13)
- Length strategy - Fixed or Variable length

```
Example features for United_States:
hasInRel_production,hasInRel_regionServed
hasInRel_citizenship,hasInRel_label
hasInRel_twin,has_percentWater
...

Example features for Washington__D_C_:
hasInRel_regionServed,has_areaTotalSqMi
has_name,has_populationTotal
hasInRel_beltwayCity,hasInRel_venue
...
```

Figure 11. Features generated for Stay Step Category with l =2

```
Example features for United_States:
office<-birthPlace->
citizenship<-placeOfDeath->
leaderName->predecessor<-
...

Example features for Washington__D_C_:
leaderTitle->order<-
stadium<-previous<-
placeOfBirth<-director<-
...
```

Figure 12. Features generated for Move Step Category with l =2

```
Example features for United_States:
hasInRel_institution,nationality<-
hasInRel_production,hasInRel_restingPlace
hasInRel_firstAired,hasInRel_releaseDate
...

Example features for Washington__D_C_:
hasInRel_governingBody,campus<-
placeOfBirth<-birthPlace->
hasInRel_residence,areaServed<-
...
```

Figure 13. Features generated for Both Step Category with l =2.

Table 6. Results of Hyper-Parameter Experimentation.

| # | Length of Walk | Step Category | Length Strategy | $F_1$-score for $n$ steps | | |
|---|---|---|---|---|---|---|
| | | | | $n=25$ | $n=50$ | $n=100$ |
| 1 | 2 | Stay | Fixed Length | 0.9144 | 0.9182 | 0.9190 |
| 2 | 2 | Stay | Variable Length | 0.9122 | 0.9168 | 0.9177 |
| 3 | 2 | Move | Fixed Length | 0.8401 | 0.8435 | 0.8456 |
| 4 | 2 | Move | Variable Length | 0.8260 | 0.8328 | 0.8393 |
| 5 | 2 | Both | Fixed Length | 0.9126 | 0.9211 | 0.9260 |
| 6 | 2 | Both | Variable Length | 0.9083 | 0.9137 | 0.9166 |
| 7 | 3 | Stay | Fixed Length | 0.8945 | 0.9037 | 0.9102 |
| 8 | 3 | Stay | Variable Length | 0.9015 | 0.9078 | 0.9091 |
| 9 | 3 | Move | Fixed Length | 0.8317 | 0.8427 | 0.8509 |
| 10 | 3 | Move | Variable Length | 0.8283 | 0.8215 | 0.8299 |
| 11 | 3 | Both | Fixed Length | 0.7474 | 0.8045 | 0.8669 |
| 12 | 3 | Both | Variable Length | 0.8742 | 0.8785 | 0.8757 |

Table 6 shows the results of the experiments for the above hyper-parameters. From these, we can conclude:

- As the number of walks and features extracted increases, the $F_1$-score also increases (consistent with previous experiments).

- Between fixed and variable length strategies, the fixed length works better, possibly due to the presence of more complex features as a result of the longer walk length.

- Use of both step categories performs best. With 2 steps, moving across a relationship creates new features that encode the type information of the adjacent node. So essentially, we are trying to identify the type of the node based on the type of the adjacent node. While the strategy to use attributes, relationship and incoming relationship presence works better than the strategy to move across an outgoing or incoming relationship, their combination works better than either of them.

- Between features with lengths 2 and 3, a feature with length 2 performs better for the selected number random walks. Though, looking at result #11, perhaps there may be a larger gain for more number of random walks.

### 5.3. Refinement of Network Structure and Random Walk Hyper-parameters

For our final set of experiments, we taper the width of the network as we go deeper. Intuitively, by doing this we are forcing the network to learn more complex representations in another dimension space (with fewer dimensions). Since this may require more learning, we choose 8 epochs for the training. We attempted to optimize our network by using Leaky ReLUs [24], but the learning was worse. Figure 14 shows the final network. In our random walk algorithm, we additionally ensured that the walk did not contain cycles. The number of walks was also increased to 125. We restrict the length of the walk to 2 and 3 with the fixed-length strategy and only use the Move and Both step categories. The last optimization we made was to select only distinct attributes and relationships in the stay step category and to move only across distinct relationships in the move step category. By doing this, we ensure that our random walk is able to extract more diverse features about the different types of neighbours.

For training the Neural Network using several batches, our dataset was split into batches of up to 5000 individuals. While the theory behind it is not completely developed, empirical observations by Keskar et. al [25] suggest that a small batch size, without going too small, leads to better performance due to smaller gradients. And so, instead of using the *train_on_batch* function in Keras, we switch to using the fit function with an inner batch of length 256 from our original batch. We put one additional restriction on the DBpedia-Categories & DBpedia-YagoTypes datasets - we only consider types and categories with at least 200 instance support. This was done since the long-tail of classes with fewer than 200 instance support becomes unmanageable for our classifier as the target vector causes an out of memory error. As such, our comparison with the DBpedia- Yago benchmarks should be taken in this changed context. The final number of classes in DBpedia-YagoTypes was 2083 and that in DBpedia-Categories was 10999. Finally, we also instead of the original train-test split, we split the dataset into training (70%), validation (20%) and test (10%). We then proceed with the final experiments for all the three datasets. The implementation details of our algorithm and relevant code are available online (https://github.com/rparundekar/understanding-things-with-semantic-graph-classification).

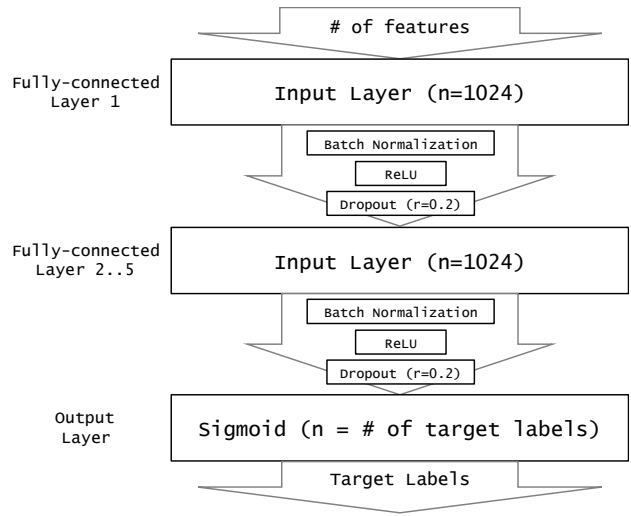

Figure 14. Final Fully-connected Network Structure

## 6. RESULTS

Table 7 shows the $F_1$-score for the validation and test splits for the DBpedia-OntologyTypes, DBpedia-Categories and DBpedia-YagoTypes datasets based on refinements above.

Table 7. Results of Type Inferencing Phase.

| # | Dataset | Step Category | # distinct features | $F_1$-score | |
|---|---------|---------------|---------------------|-------------|---|
|   |         |               |                     | *Validation* | *Test* |
| 13 | DBpedia-OntologyTypes | Move | 159,766 | 0.8668 | 0.8668 |
| 14 | DBpedia-OntologyTypes | Both | 233,419 | 0.9203 | 0.9200 |
| 15 | DBpedia-YagoTypes | Move | 167,357 | 0.8576 | 0.8594 |
| 16 | DBpedia-YagoTypes | Both | 729,468 | 0.8550 | 0.8549 |
| 17 | DBpedia-Categories | Move | 73,442 | 0.2909 | 0.2895 |
| 18 | DBpedia-Categories | Both | 390,373 | 0.3388 | 0.3392 |

As noted by both SDtype and SLCN, in the original generation of DBpedia from Wikipedia, the types of the individuals were assigned using the infobox properties and the attributes and relationships were extracted for the identified types. Based on this, the authors did not use attributes and outgoing relationships for their classification citing triviality. And so, when comparing with these below, we choose the results of our algorithm with features generated using the Move strategy having walk length greater than one.

Table 8. Comparison with Prior-Art.

|   | Dataset | SDtype | SLCN | *This Work (Move, l=2, Fixed and n=125)* | *Gold Standard (all attributes)* |
|---|---|---|---|---|---|
| A | DBpedia-OntologyTypes | 0.765 | 0.847 | 0.8668 | 0.9254 |
| C | DBpedia-YagoTypes | 0.666 | 0.702 | 0.8594 | N/A |

For DBpedia-OntologyTypes, SDtype achieves an $F_1$-score of 0.765 and SLCN achieves a score of 0.847. In comparison, our model with random walks of fixed length 2 and moving to an adjacent individual (i.e. using Move step category) performs better and achieves a $F_1$-score of 0.8668. Compared to using all outgoing relationships with random walk of length one as shown in Table 3, our refinements and longer walks also resulted in increase of the $F_1$-score from 0.8025 to 0.8668. For the DBpedia-YagoTypes dataset, SDtype achieves a $F_1$-score of 0.666 and SLCN achieves a score of 0.702. In comparison, our model with random walks of fixed length 2 and using Move step category achieves a $F_1$-score of 0.8594. Again, our model beats the two benchmarks. However, since we remove the long-tail types with a support of fewer than 200 individuals, this needs further confirmation. For DBpedia-Categories dataset, our models trained on both Move as well as Both step categories achieve a $F_1$-score of 0.2895 and 0.3392 respectively (Note: no prior-art available for comparison).

We also consider the accuracy of the final Deep Neural Network trained with all 3 steps of attribute, relationship, and incoming relationship presence types as a gold standard for the DBpedia-OntologyTypes dataset. To generate this, we re-trained the final Deep Neural Network on the dataset containing 4169 features extracted with all three step types similar to Table 5. With 15 epochs of training, we achieved a maximum $F_1$-score of 0.9254. Both models trained on Move as well as Both step categories with length 2 achieve comparable performance of a $F_1$-score of 0.9200 for this dataset.

While in our initial experiments, relying on all incoming relationship presence features only achieved an $F_1$-score of 0.7596 for 8-layer Fully-connected Neural Network (see Table 4), SLCN achieves an impressive 0.847. Since our approach with an $F_1$-score of 0.8668 uses outgoing links in the random walks of fixed length 2 with Move step category, which SLCN avoids because they are generated from the type information, it is not completely fair to conclude that our approach performs better. At the same time, the features generated by the random walk create a neighbourhood context since they also consider the relationships of the adjacent nodes, which is worthwhile to compare with SLCN. And so, we conclude with the above caveat that our approach combining feature extraction using random walks and multi-label classification using Deep Neural Networks can be very effective in extracting contexts from neighbourhoods of individuals in a semantic graph and present a promising alternative to state-of-the-art methods.

Since our approach is able to perform classification in noisy data and also works on an unseen validation set with the same performance as train and test datasets, we can reasonably say our approach and chosen hyper-parameters make our classifier robust. While we only investigate DBpedia due to limited dataset availability for comparison with prior art, our approach is extendable to any other semantic dataset as it only relies on the symbolic names associated with attributes and relationships to extract features using random walks and is thus generalizable.

## 6. CONCLUSION AND FUTURE WORK

The main contribution of this work is in creating a novel approach for robust multi-label classification by extracting features from a semantic graph such as DBpedia and using Deep Neural Networks to identify types of the individuals in the presence of noisy data on the Semantic Web. In our two-step approach, we first extract features by performing random walks on the individuals to be classified, and then perform multi-label classification using a Deep Neural Network. We identified the effect of hyper-parameters like length and number of random walks, the type of steps taken, the fully-connected Neural Network architecture, etc. on the classification performance through systematic evaluation. After selecting the best hyper-parameters based on these experiments, we evaluated the performance of our solution and compared it with prior-art. The results show that our method consistently performs better than systems like SDtype and SLCN, from which we can conclude that this approach can be very effective in extracting contexts from neighbourhoods of individuals in a semantic graph and presents a promising alternative to state-of-the-art methods.

There are two possible areas of future work. First, while our approach is robust for noise in the RDF data on the Semantic Web, we need to investigate label de-noising techniques such as Wang et. al [26], Imani et. al [27], etc. after a more detailed label analysis by using techniques such as Xie et. al [28]. Second, we feel that our approach can indeed outperform prior-art, but needs a more rigorous experimentation and training hyper-parameter tuning of both the feature extraction using random walks as well as the Deep Neural Network steps. While more work is needed on confirming our finding with other datasets and with benchmark datasets, our approach seems to be generalizable to understanding the meaning and context of things in any Semantic Graph not just limited to the Semantic Web (for example, those mentioned in the Section 1).

## REFERENCES

[1] T. Berners-Lee, J. Hendler, O. Lassila et al., "The semantic web," Scientific american, vol. 284, no. 5, pp. 28–37, 2001.

[2] T. Heath and C. Bizer, "Linked data: Evolving the web into a global data space," Synthesis lectures on the semantic web: theory and technology, vol. 1, no. 1, pp. 1–136, 2011.

[3] R. Socher, C. C. Lin, C. Manning, and A. Y. Ng, "Parsing natural scenes and natural language with recursive neural networks," in Proceedings of the 28th international conference on machine learning (ICML-11), 2011, pp. 129–136.

[4] L. Backstrom and J. Leskovec, "Supervised random walks: predicting and recommending links in social networks," in Proceedings of the fourth ACM international conference on Web search and data mining. ACM, 2011, pp. 635–644.

[5] J.-L. Lugrin and M. Cavazza, "Making sense of virtual environments: action representation, grounding and common sense," in Proceedings of the 12th international conference on Intelligent user interfaces. ACM, 2007, pp. 225–234.

[6] I. Horrocks, P. F. Patel-Schneider, and F. Van Harmelen, "From shiq and rdf to owl: The making of a web ontology language," Web semantics: science, services and agents on the World Wide Web, vol. 1, no. 1, pp. 7–26, 2003.

[7] H. Paulheim and C. Bizer, "Type inference on noisy rdf data," in International Semantic Web Conference. Springer, 2013, pp. 510–525.

[8] A. Hogan, A. Harth, J. Umbrich, S. Kinsella, A. Polleres, and S. Decker, "Searching and browsing linked data with swse: The semantic web search engine," Web semantics: science, services and agents on the world wide web, vol. 9, no. 4, pp. 365–401, 2011.

[9] B. Tang, "The emergence of artificial intelligence in the home: Products, services, and broader developments of consumer oriented ai," 2017.

[10] A. Schmeil and W. Broll, "Mara-a mobile augmented reality-based virtual assistant," in Virtual Reality Conference, 2007. VR'07. IEEE. IEEE, 2007, pp. 267–270.

[11] M. Cordts, M. Omran, S. Ramos, T. Scharwächter, M. Enzweiler, R. Benenson, U. Franke, S. Roth, and B. Schiele, "The cityscapes dataset," in CVPR Workshop on the Future of Datasets in Vision, vol. 1, no. 2, 2015, p. 3.

[12] H. Paulheim, "Knowledge graph refinement: A survey of approaches and evaluation methods," Semantic web, vol. 8, no. 3, pp. 489–508, 2017.

[13] Q. Miao, R. Fang, S. Song, Z. Zheng, L. Fang, Y. Meng, and J. Sun, "Automatic identifying entity type in linked data," PACLIC 30, p. 383, 2016.

[14] A. Melo, J. Völker, and H. Paulheim, "Type prediction in noisy rdf knowledge bases using hierarchical multilabel classification with graph and latent features," International Journal on Artificial Intelligence Tools, vol. 26, no. 02, p. 1760011, 2017.

[15] A. Clare and R. King, "Knowledge discovery in multi-label phenotype data," Principles of data mining and knowledge discovery, pp. 42–53, 2001.

[16] K. Tsuda and H. Saigo, "Graph classification," Managing and mining graph data, pp. 337–363, 2010.

[17] G. Tsoumakas and I. Katakis, "Multi-label classification: An overview," International Journal of Data Warehousing and Mining, vol. 3, no. 3, 2006.

[18] J. Nam, J. Kim, E. L. Mencía, I. Gurevych, and J. Fürnkranz, "Largescale multi-label text classificationrevisiting neural networks," in Joint european conference on machine learning and knowledge discovery in databases. Springer, 2014, pp. 437–452.

[19] B. Perozzi, R. Al-Rfou, and S. Skiena, "Deepwalk: Online learning of social representations," in Proceedings of the 20th ACM SIGKDD international conference on Knowledge discovery and data mining. ACM, 2014, pp. 701–710.

[20] J. Lehmann, R. Isele, M. Jakob, A. Jentzsch, D. Kontokostas, P. N. Mendes, S. Hellmann, M. Morsey, P. Van Kleef, S. Auer et al., "Dbpedia–a large-scale, multilingual knowledge base extracted from wikipedia," Semantic Web, vol. 6, no. 2, pp. 167–195, 2015.

[21] F. M. Suchanek, G. Kasneci, and G. Weikum, "Yago: a core of semantic knowledge," in Proceedings of the 16th international conference on World Wide Web. ACM, 2007, pp. 697–706.

[22] X. Glorot and Y. Bengio, "Understanding the difficulty of training deep feedforward neural networks." in Aistats, vol. 9, 2010, pp. 249–256.

[23] T. Dozat, "Incorporating nesterov momentum into adam," Stanford University, Tech. Rep., 2015. [Online]. Available: http://cs229. stanford. edu/proj2015/054 report. pdf, Tech. Rep., 2015.

[24] A. L. Maas, A. Y. Hannun, and A. Y. Ng, "Rectifier nonlinearities improve neural network acoustic models," in Proc. ICML, vol. 30, no. 1, 2013.

[25] N. S. Keskar, D. Mudigere, J. Nocedal, M. Smelyanskiy, and P. T. P. Tang, "On large-batch training for deep learning: Generalization gap and sharp minima," arXiv preprint arXiv:1609.04836, 2016.

[26] D. Wang and X. Tan, "Label-denoising auto-encoder for classification with inaccurate supervision information," in Pattern Recognition (ICPR), 2014 22nd International Conference on. IEEE, 2014, pp. 3648–3653.

[27] M. Imani, and U.M. Braga-Neto, "Control of Gene Regulatory Networks with Noisy Measurements and Uncertain Inputs," IEEE Transactions on Control of Network Systems (TCNS), 2018.(DOI: 10.1109/TCNS.2017.2746341)

[28] S. Xie, et. al. "Nonstationary Linear Discriminant Analysis", In 51th Asilomar Conference on Signals, Systems and Computers. IEEE 2017.

## AUTHORS

Rahul Parundekar completed his Masters in Computer Science from the University of Southern California, Los Angeles in 2010 and has been working as a Senior Researcher investigating Intelligent Assistants in the Automotive Industry.

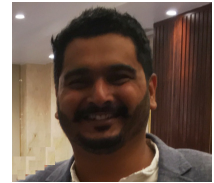